\documentclass[letterpaper, 10 pt, conference]{ieeeconf}  

\IEEEoverridecommandlockouts                              

\title{\LARGE \bf
	A Fully Automated System for Sizing Nasal PAP Masks Using Facial Photographs}

\author{Benjamin Johnston \textit{Student Member, IEEE} and Philip de Chazal \textit{Senior Member, IEEE}
	\thanks{Research supported by ARC grant FT110101098}
	\thanks{Ben Johnston and Philip de Chazal are with the Sleep Research Group, Charles Perkins Centre, School of Electrical and Information Engineering, University of Sydney, Sydney, NSW, 2006, Australia, (phone: +612 911 41528; e-mails: \tt\small{\{ben.johnston, philip.dechazal\}@sydney.edu.au)}}%
}
\usepackage{amsmath, graphicx, wrapfig, verbatim, amssymb, wasysym, multirow}

\begin{document}
	%
	\maketitle
	\begin{abstract}
		We present a fully automated system for sizing nasal Positive Airway Pressure (PAP) masks.  The system is comprised of a mix of HOG object detectors as well as multiple convolutional neural network stages for facial landmark detection.  The models were trained using samples from the publicly available PUT and MUCT datasets while transfer learning was also employed to improve the performance of the models on facial photographs of actual PAP mask users.  The fully automated system demonstrated an overall accuracy of 64.71\% in correctly selecting the appropriate mask size and 86.1\% accuracy sizing within 1 mask size. 
		
	\end{abstract}
	\section{Introduction}
	\label{sec:intro}
	Sleep apnoea is a condition which leads to the repetitive cessation of breathing during sleep and commonly produces symptoms such as poor quality sleep, excessive daytime sleepiness, headaches and weight gain, as well as being associated with an elevated risk of other diseases such as cardiovascular disease and diabetes \cite{Weaver2011}.  The most common form of sleep apnoea Obstructive Sleep Apnoea (OSA) is caused by either the partial or complete blockage of the upper airway  and can be effectively treated by preventing the blockage from occurring.  Since its inception in 1981 by Sullivan et al. \cite{Sullivan1981},  positive airway pressure, or continuous positive airway pressure (CPAP) has been the gold standard for the treatment of OSA.  PAP therapy is applied through the use of a specialised medical device which delivers a highly controlled wave of pressurised air to the upper airway, acting as a pneumatic splint and preventing the blockage of the pharynx that characterises OSA.  One vital component of the CPAP device is the specialised mask that must be worn by the patients during sleep.  The mask must be able to provide a seal that maintains the required therapy pressure to ensure airway remains open, despite any movement of the patient's head or body or collisions of the mask with the bed or pillow.  The mask must also be sufficiently comfortable to allow a patient to wear it during hours of sleep, which is particularly important as while maintenance of therapy pressure is essential for treatment, mask related side effects have been shown to dissuade patients from actually using their device \cite{Massie2003}.  Patients who do not use their device obviously do not receive any benefit from therapy.  Mask related side effects such as red marks, pressure leakage into the eyes and skin abrasions have been reported in previous studies \cite{Berry2000, Gay2006} with one such study reporting the occurrence of mask leak and red marks in 48\% and 40\% of participants respectively \cite{Amfilochiou2009}.  Given that mask related performance issues are a significant contributor to patient usage rates and rates of discontinuing therapy \cite{Engleman1994} it is extremely important the mask issued to a patient is the most appropriate model and size.
	
	\subsection{Current Mask Sizing Methods}
	\label{sec:current_sizing}
	
\begin{wrapfigure}{l}{.48\linewidth}
			\centering
			\centerline{\includegraphics[height=3.5cm]{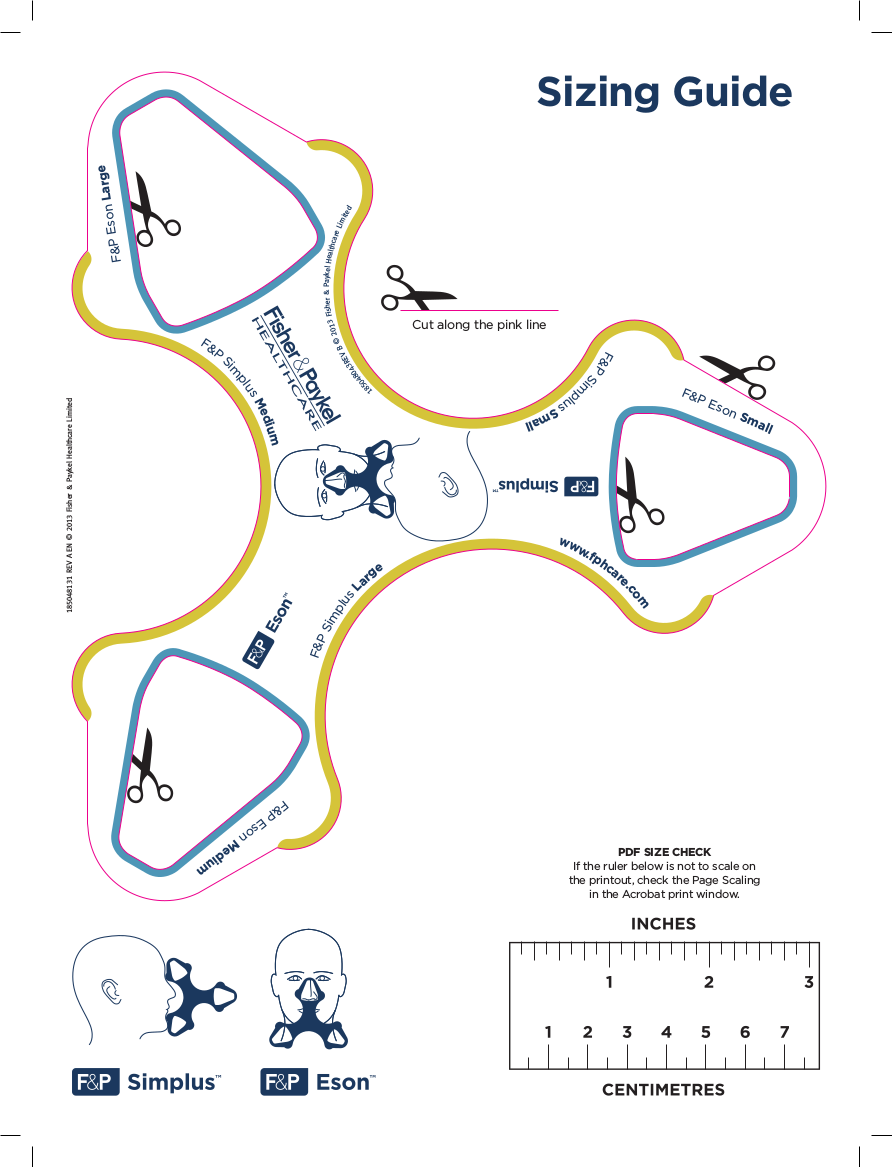}}
			\caption{ESON Mask Sizing Guide}
		\label{fig:size_template}
	\end{wrapfigure}

	The exact process for selecting, sizing and issuing CPAP masks varies depending upon regulatory jurisdictions, manufacturers requirements and clinician's own professional experience.  In general however the process for selecting the most appropriate mask size can be summarised into one of two methods: use of a manufacturer supplied sizing template and use of previous experience and expertise. Almost every manufacturer provides a physical guide for determining the most appropriate mask size for a patient.  These guides (such as that shown in Figure \ref{fig:size_template}) use the position of patients nares relative to specific markers as a means of indicating the correct mask size.  Thus, the ESON nasal mask sizes are as follows: Small: 0 - 37mm, Medium: 37 - 41mm, Large: 41 - 45mm and Too Large $>$ 45mm.  While Too Large is not an official mask size, it is used to indicate if a patient would not fit any of the mask size options.

	The second and most commonly used method employs the experience and expertise of the clinicians as well as the feedback of the patient during the consultation process, without the use of the abovementioned guide.  This method, while not always issuing the manufacturer recommended size can easily and quickly incorporate in-field feedback during the decision making process.
	
	As the cost of healthcare across the world continues to increase, and with an estimated 59\% of the adult population suffering from sleep apnoea \cite{Ulualp2010, JM2011}, there is an increasing need to ensure that the time clinicians spend with patients is most effectively used.  The use of a fully automated mask sizing system such as that described within this article could assist in alleviating some of this time pressure for both the clinician and patient; while combined with other telemedicine services could allow consultations to be completed remotely, removing the need for remote and rural patients to travel the often large distances to their nearest clinic.
	
	The work presented within this paper extends upon our previous work where a semi-automated solution for sizing nasal PAP masks was presented\cite{Johnston2017}; still requiring human intervention to determine the scale of the image being presented.  The solution presented within this paper is one step closer to assisting clinicians in that human intervention is not required to predict the appropriate mask size.
	
	\section{Datasets}
	\label{sec:datasets}
	
	A total of four separate datasets were used during the development of the automated CPAP mask sizing system, including two publicly available datasets: MUCT \cite{Milborrow10} and PUT \cite{Kasinski2008}.  One dataset collected during this study contains 206 facial photographs of male and female subjects, many of which were PAP patients with an Australian 20 cent piece positioned on their forehead.  Physical nose width measurements were also taken of the participants using vernier calipers; henceworth this dataset will be referred to as the \textit{test set}.  One additional dataset of 131 facial photographs of PAP and non-PAP participants was collected without any physical measurements or the presence of the coin in the image; this dataset will be referred to as the \textit{post-train set}.  One final dataset, described in more detail below was artificially generated using both the MUCT and PUT datasets in addition to photographs of both the head and tails side of an Australian 20 cent piece, downloaded from the internet.
	
	\begin{figure}[h]
		\begin{minipage}[b]{.48\linewidth}
			\centering
			\centerline{\includegraphics[height=1.5cm]{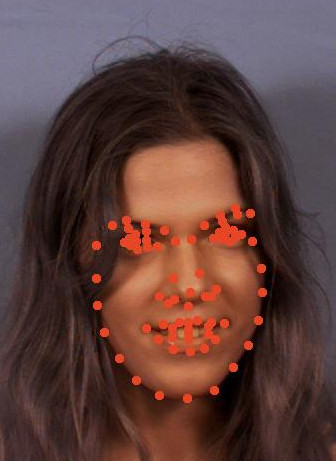}}
			
			\centerline{(a) MUCT example}\medskip
			
		\end{minipage}
		\hfill
		\begin{minipage}[b]{0.48\linewidth}
			\centering
			\centerline{\includegraphics[height=1.5cm]{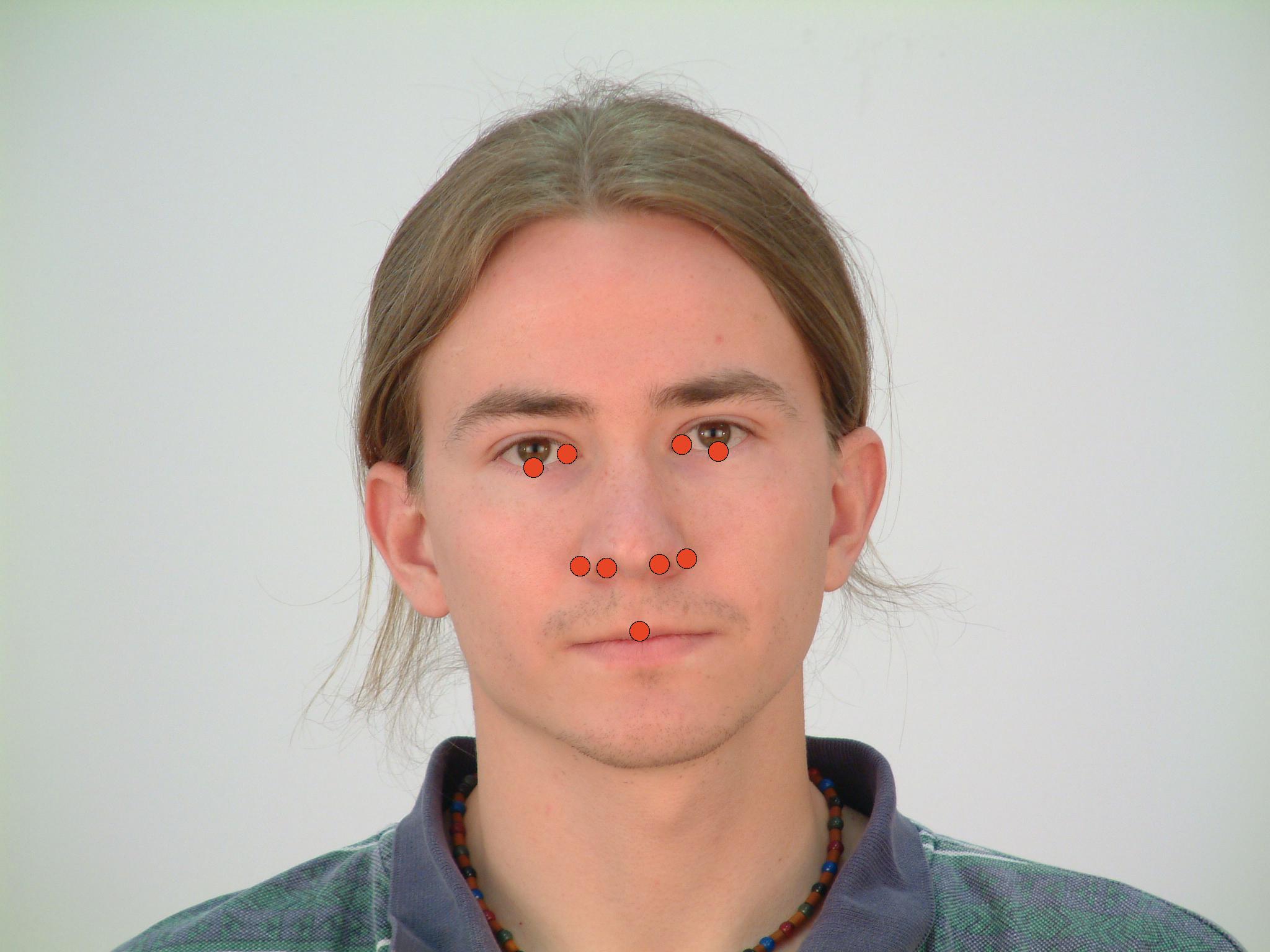}}
			\centerline{(b) PUT example}\medskip
		\end{minipage}		
		
		\begin{minipage}[b]{.48\linewidth}
			\centering
			\centerline{\includegraphics[height=1.5cm]{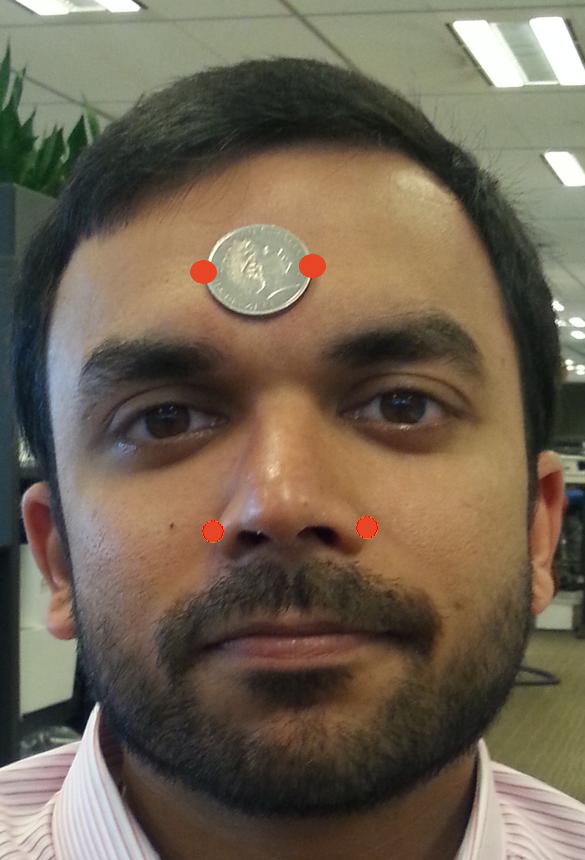}}
			
			\centerline{(c) \textit{Test-set} \#}\medskip
			
		\end{minipage}
		\hfill
		\begin{minipage}[b]{0.48\linewidth}
			\centering
			\centerline{\includegraphics[height=1.5cm]{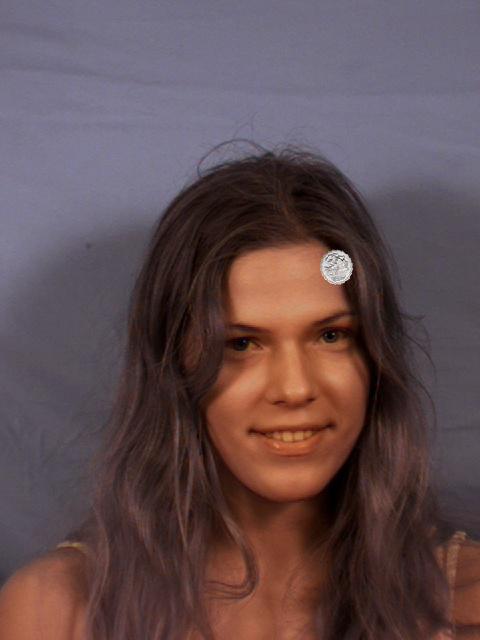}}
			\centerline{(d) Generated coin data}\medskip
		\end{minipage}
		
		\caption{Datasets (\# Image used with permission from Priyanshu Gupta)}
		\label{fig:datasample}
	\end{figure}

    Samples from the MUCT and PUT datasets were selected for use during this study, as opposed to others databases such as 300W \cite{Sagonas2016} and menpo \cite{Zafeiriou_2017_CVPR_Workshops} due to their similarity to the data collected during this study.  The MUCT dataset, while containing images of approximately the same pose, with different lighting is known for being composed of participants from a variety of demographics; with both sexes, a range of different ages and ethnicities being represented in the set.  The MUCT samples selected for use where those images captured from camera angle $a$ with all lighting options, both genders and participants with and without glasses. 
    
    The PUT dataset, while not as ethnically diverse as the MUCT dataset, contains samples of varied lighting and approximately the same pose as that shown in the datasets containing PAP patients.  All samples in the PUT dataset were selected for use.  Another reason for the selection of both MUCT and PUT is that these datasets contain landmark annotations identifying the width of the nose as required by the nasal PAP mask sizing guide.  These landmarks were manually annotated for images taken during this study and for the test set, landmarks indicating the width of the coin were also manually annotated.
    
    \subsection{Artificially Generated Coin Dataset}
    \label{sec:art_coin_data}
    
    Given that only 245 samples of photographs of Australian 20 cent pieces were available for use it was decided that more data would be required to produce models capable of detecting the width of the coin with sufficient accuracy and allowing a large enough test set to measure system performance.  To overcome this hurdle and remove the need to take a large number of images of a coin, the MUCT and PUT datasets were used to artificially generate images of the coin similar to that present within the test set.  Six images of the heads side of a 20c piece were downloaded from the internet in addition to another six images of the standard tails side of coin containing impressions of the platypus design.

    For each image in the sample of MUCT and PUT datasets, one of the 12 coin images were randomly sampled for inclusion in the generated sample.  For the MUCT dataset the intra-ocular distance was used to scale the image of the coin to an appropriate size for the subject's face; for the PUT samples the distance between the inner corners of the eye was used.  The approximate location of the coin in the image was specified to be above the highest landmark in the set and inline with the nose tip.  Using uniform random sampling, an offset was then applied to this position to increase the variability of the coin position with respect to the face.  The image of the coin was then rotated by a random angle between -60 and 60 degrees to the horizontal, a random gaussian blur was applied with $1 < \sigma < 4$ and finally the brightness and contrast was randomly adjusted within $\pm20\%$.  All random values were generated using uniform random sampling.
    
    \section{Methods}
    \label{sec:methods}
    
\begin{figure*}[!htb]
    	
    	\begin{minipage}[t]{1\linewidth}
    		\centering
    		\centerline{\includegraphics[height=2.5cm]{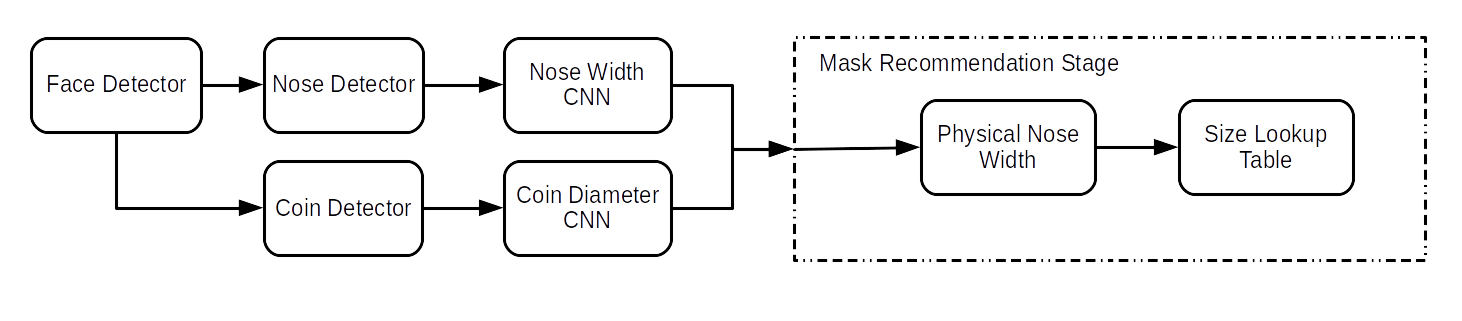}}
    	\end{minipage}
    	\caption{System Architecture}
    	\label{fig:sys_arch}
    \end{figure*}

    The system architecture, as shown in Figure \ref{fig:sys_arch} is composed of multiple object detectors (face, nose and coin), convolutional neural networks for landmark identification followed by a stage which combines the results of landmark identification stages to compute a recommended mask size.  The face detector is the entry point for the system, as all information that is required by the system is contained within the bounding box of the face.  In this implementation the OpenCV implementation of the Viola \& Jones algorithm \cite{Viola2001} was used with the pre-trained front-on face detector.  For the nose and coin detection stages; the Dlib implementation \cite{King2009} of the Dalal \& Triggs HOG based detector \cite{Dalal2005} was used, training the detectors using a subset of the noses present within the MUCT and PUT datasets as well as a sample of the coins within the generated dataset.
    
    \subsection{Nose Width CNN Stage}
    \label{sec:nose_width_cnn}
    
    After the nasal region is extracted by the nose detector the sub-image and corresponding training landmarks were resized to
    42 x 42 RGB pixels, while maintaining the perspective of the image.  Those images, which after resizing had a height of less than 42 pixels, were filled with values of zero to reach the required size.  For each training sample, the matrix of $2\times2$ landmarks were flattened to form a single 4 unit vector.  The image and coordinate values were then scaled between -1 and 1 by dividing by their respective maxima and subtracting the mean; when applying the test set to the model, only the input image values required scaling.
    
    The nasal regions of each image of the MUCT/PUT datasets, along with the nose width coordinates were first extracted, then left/right flipped copies of the data were  constructed to double the amount of data available for training.  The selected data was randomly split into training and validation sets with a consistently seeded random number generator; 70\% of the data was allocated for training and 30\% for validation.  The training and validation sets were then used to tune the design of the nose width model, starting with a simpler, single hidden layer neural network model and increasing the number of hidden units, number of layers, as well as number of convolutional and pooling layers until the error score was minimised.  A \textit{relu} non-linearity function was applied to hidden and convolutional layers, max pooling was used for all pooling layers and a root mean square error function applied as the objective function.  All weights, including weights for bias values were initialised with random values between $\pm \sqrt{6 / (units_{in} + units_{out})}$; stochastic gradient descent was applied using minibatches of 128 samples, a fixed learning rate of $1\times10^{-4}$, momentum of 0.9 and patience of 50 epochs.  After the hyperparameter sweep the following architecture was selected: 1 Convolutional Layer with a $3\times3$ filter and 32 feature maps, a max-pooling layer a fully connected layer with 800 hidden units and the output layer with 4 units.
    
    Following training, the nose detector was then applied to 83 facial photographs of PAP patients which along with the manually annotated landmarks were normalised and reshaped to continue the training process.  The samples were also flipped left to right to create an additional 83 samples for training.  The previously saved weights were reloaded into the model and stochastic gradient descent continued using a batch size of 1 and instead of a 70\%/30\% training / validation set split, 90\% of the data was allocated for training.  The significantly larger training set was specified to provide the model with as many samples as possible to learn the features of the actual data of interest.
    
    \subsection{Coin CNN Stage}
    \label{sec:coin_CNN}
    
    The coin CNN stage is the means by which the automated system determines the scale of the facial features within the image; knowing that the diameter of an Australian 20 cent piece is $28.65$mm, by detecting the width of the coin in the image, the scale in pixels per millimetre (px/mm) was determined.  Following the application of the HOG coin detector, the resulting sub-image is resized and padded using the same process as used during the nose CNN stage, to be consistent with the smallest available image i.e. 42 x 42 RGB pixels.  The resulting image is then normalised as previously described.  
    
    During training of the coin CNN stage, the artificially generated dataset was also split with 70\% of the data allocated for training.  The same hyperparameter selection process as was completed with the nose width CNN stage was repeated.  After the hyperparameter sweep, the following architecture was selected: 1 Convolutional Layer with a $3\times3$ filter and 24 feature maps, a max-pooling layer, a fully connected layer with 800 hidden units and the linear output layer with 4 units.
    
    Once the training error had been minimised using the artificially generated data, transfer learning was again used to refine the model for its intended use.  A total of 51 samples were selected from the test set for this purpose and again a 90\%/10\% training / validation set split was used. The model was loaded with the optimised weights and the gradient descent process was continued, with the 51 isolated test set samples and a batch size of one, until the error function was again minimised.

    \subsection{Mask Recommendation Stage}
    \label{sec:mask_rec}
    
    The mask recommendation stage is where the nose width measurement in pixels and the scale information as provided by the coin are combined to compute the physical nose width of the patient in millimetres.  This measurement was then compared to each of the corresponding mask sizes.  As described in section \ref{sec:current_sizing} the issuance of PAP mask sizes is not an exact science; as such PAP mask manufacturers provide an overlap for mask sizes, allowing patients who fall within an overlap to be considered correctly sized if they receive one of either of the adjacent sizes. To reflect this situation within the field we applied a 5\% tolerance to sizing decisions.  For any individual whose `ground truth' size is within 5\% of a sizing boundary, the predict size was considered correct if one of either the adjacent sizes were issued.
    
    \section{Results}
    \label{sec:results}
    
\begin{figure}[h]
		\begin{minipage}[t]{1\linewidth}
			\centering
			\centerline{\includegraphics[height=1.5cm]{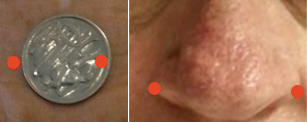}}
			
			\centerline{(a) Good Examples}\medskip
			
		\end{minipage}
		\hfill
		\begin{minipage}[b]{1\linewidth}
			\centering
			\centerline{\includegraphics[height=1.5cm]{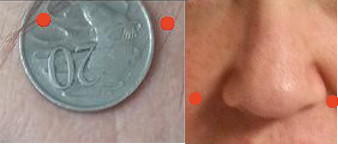}}
			\centerline{(b) Poor Examples}\medskip
		\end{minipage}		
		
		\caption{Predicted landmark results}
		\label{fig:pred_results}
	\end{figure}

    The face and nose detectors within the system correctly detected all faces and noses within the test set, while the coin detector successfully identified all but one of the present coins.  Tables \ref{tab:pred_confusion_matrix} and \ref{tab:sensitivity_predict_perf} summarise the sizing performance of the automated system, while figure \ref{fig:pred_results} provides examples of the CNN stages performing well and poorly.  Using the confusion matrix (Table \ref{tab:pred_confusion_matrix}) the fully automated PAP mask sizing system produced an accuracy of 
    64.71\% and 86.1\% within one size.

    \begin{table}[!t]
    	\centering
    	\renewcommand{\arraystretch}{1.4}
    	\begin{tabular}{cc  c | c | c | c |}
    		
    		& & \multicolumn{4}{c }{\textbf{Predicted Mask Sizes}} \\
    		\cline{3-6}
    		\multirow{6}{*}{\rotatebox[origin=l]{90}{\textbf{Actual Mask Size}}} & & \multicolumn{1}{|c |}{Small} & \multicolumn{1}{c |}{Medium} & \multicolumn{1}{c |}{Large} & \multicolumn{1}{c |}{Too Large}\\
    		\cline{2-6}
    		& \multicolumn{1}{|c |}{Small} & 75 & 15 & 5 & 8\\
    		\cline{2-6}
    		\multicolumn{1}{ c}{}
    		& \multicolumn{1}{|c |}{Medium} & 16 & 33 & 5 & 9\\
    		\cline{2-6}
    		\multicolumn{1}{ c}{}
    		& \multicolumn{1}{|c |}{Large} & 4 & 3 & 10 & 1\\
    		\cline{2-6}
    		\multicolumn{1}{ c}{}
    		& \multicolumn{1}{|c |}{Too Large} & 0 & 0 & 0 & 3\\
    		\cline{2-6}
    	\end{tabular}
    	\caption{Confusion matrix of predicted mask sizes}
    	\label{tab:pred_confusion_matrix}
    	
    	\begin{tabular}{l |c|c|c|c|c|c|c|c|}
			\cline{2-9}
			& \multicolumn{4}{|c}{\textbf{Sensitivity (\%)}} & \multicolumn{4}{|c|}{\textbf{Pos. Predict. (\%)}} \\
			\cline{2-9}
			& S & M & L & TL & S & M & L & TL\\
			\hline
			\multicolumn{1}{|c|}{\rotatebox[origin=c]{90}{\textbf{\thinspace Pred \thinspace}}} & 76 & 52 & 56 & 100 & 79 & 65 & 50 & 14\\
			\hline
		\end{tabular}
		\caption{Sensitivity \& positive predictivity performance of predicted mask sizes (S: Small, M: Medium, L: Large, TL: Too Large)}
		\label{tab:sensitivity_predict_perf}    	
    \end{table}

	\section{Discussion}
	\label{sec:discussion}
	
	The work completed throughout this study provides an excellent foundation for creating a fully automated PAP mask sizing system.  Despite the limited availability of custom data, the use of similar, publicly available datasets and transfer learning produced a system with 64.71\% accuracy on unseen data from the custom set.  In order for a similar system to be deployed for use in the field, the accuracy must be further improved.   It is expected that by increasing the size and quality of the PAP patient data set that overall performance would increase as training could occur on the actual data as opposed to `similar' data.  The dataset could also be enhanced by improving the consistency of pose, facial expressions and lighting via clearer user instructions.

	\section{Conclusion}
	\label{sec:conclusion}
	
	This study presented a fully automated method of sizing the ESON nasal PAP mask through the use of a multi stage convolutional neural network and transfer learning.  The system produced an accuracy of 
	64.71\% and 86.1\% within one size.

	\bibliographystyle{IEEEbib}
	\bibliography{refs}
	
\end{document}